\documentclass[lettersize,journal]{IEEEtran}
\usepackage{hyperref}
\usepackage{amsmath,amssymb,amsfonts}
\usepackage{graphicx}
\usepackage{textcomp}
\usepackage{algorithm,algcompatible}
\usepackage{makecell} 
\usepackage{algpseudocode}
\usepackage{graphicx}
\usepackage{booktabs}
\usepackage{xspace}
\usepackage[table,xcdraw]{xcolor}
\usepackage{tikz}
\usetikzlibrary{patterns}
\usepackage{pgfplots}
\usepackage{pgfplotstable}
\usepackage{newfloat}
\usepackage{listings}
\pgfplotsset{compat=1.18}
\usepackage{amsmath}
\usetikzlibrary{arrows.meta,
                chains,
                positioning}
\usepackage{subcaption}
\begin{document}

\title{Adaptive Multi-Agent Deep Reinforcement Learning for Timely Healthcare Interventions}

\author{Thanveer Shaik,  Xiaohui Tao, Lin Li, Haoran Xie,  Hong-Ning Dai, Feng Zhao and Jianming Yong
\thanks{Thanveer Shaik and Xiaohui Tao are with 
the School of Mathematics, Physics \& Computing, University of Southern Queensland, Queensland, Australia (e-mail: Thanveer.Shaik@unisq.edu.au, Xiaohui.Tao@unisq.edu.au).}
\thanks{Lin Li is with the School of Computer and Artificial Intelligence, Wuhan University of Technology, China (e-mail: cathylilin@whut.edu.cn)}
\thanks{Haoran Xie is with the Department of Computing and Decision Sciences, Lingnan University, Tuen Mun, Hong Kong (e-mail: hrxie@ln.edu.hk)}
\thanks{Hong-Ning Dai is with the Department of Computer Science, Hong Kong Baptist University, Hong Kong (e-mail: henrydai@hkbu.edu.hk).}
\thanks{Feng Zhao  is with Huazhong University of Science and Technology, Wuhan, China (email: zhaof@hust.edu.cn).}
\thanks{Jianming Yong is with the School of Business, University of Southern Queensland,  (e-mail: Jianming.Yong@unisq.edu.au)}
}

\markboth{Journal of \LaTeX\ Class Files,~Vol.~14, No.~8, August~2021}%
{Shell \MakeLowercase{\textit{et al.}}: A Sample Article Using IEEEtran.cls for IEEE Journals}


\maketitle

\begin{abstract}
Effective patient monitoring is vital for timely interventions and improved healthcare outcomes. Traditional monitoring systems often struggle to handle complex, dynamic environments with fluctuating vital signs, leading to delays in identifying critical conditions. To address this challenge, we propose a novel AI-driven patient monitoring framework using multi-agent deep reinforcement learning (DRL). Our approach deploys multiple learning agents, each dedicated to monitoring a specific physiological feature, such as heart rate, respiration, and temperature. These agents interact with a generic healthcare monitoring environment, learn the patients' behavior patterns, and make informed decisions to alert the corresponding Medical Emergency Teams (METs) based on the level of emergency estimated. In this study, we evaluate the performance of the proposed multi-agent DRL framework using real-world physiological and motion data from two datasets: PPG-DaLiA and WESAD. We compare the results with several baseline models, including Q-Learning, PPO, Actor-Critic, Double DQN, and DDPG, as well as monitoring frameworks like WISEML and CA-MAQL. Our experiments demonstrate that the proposed DRL approach outperforms all other baseline models, achieving more accurate monitoring of patient's vital signs. Furthermore, we conduct hyperparameter optimization to fine-tune the learning process of each agent. By optimizing hyperparameters, we enhance the learning rate and discount factor, thereby improving the agents' overall performance in monitoring patient health status. Our AI-driven patient monitoring system offers several advantages over traditional methods, including the ability to handle complex and uncertain environments, adapt to varying patient conditions, and make real-time decisions without external supervision. However, we identify limitations related to data scale and prediction of future vital signs, paving the way for future research directions.
\end{abstract}


\section{Introduction}\label{sec:introduction}

In the dynamic domain of healthcare, the significance of informed decision-making cannot be overstated. With the advent of continuous patient monitoring systems, it has become possible to remotely track vital signs and physical movements, thereby enhancing the decision-making capabilities of clinicians~\cite{el2021mobile, shaik2022ai}. The application of machine learning models to analyze transmitted vital sign data has seen a significant uptick in various healthcare applications, ranging from pre-clinical data processing and diagnosis assistance to early warning detection of health deterioration, treatment decision-making, and drug prescription~\cite{Rana2022, bohr2020rise}. In this context, the monitoring of human behavior patterns plays a crucial role, especially for remote patient monitoring in hospitals or through Internet of Things (IoT)-enabled home monitoring systems.

Traditionally, methodologies in this field have predominantly relied on unsupervised and supervised learning techniques to identify patterns and classify patients' activities and vital signs~\cite{Pattanayak2022,Thirunavukarasu2022}. However, these techniques are limited in their capacity to only observe data and suggest potential decisions without the ability to act upon these observations. In contrast, Reinforcement Learning (RL) introduces a novel paradigm by deploying learning agents within complex and uncertain environments. These agents are empowered to explore and exploit the environment through actions, learning from the outcomes of their actions~\cite{Kiran2022}. A cornerstone of RL is its reward mechanism, which provides the agent with feedback in the form of rewards for its actions. These rewards serve as crucial signals that guide the learning process of the agent, encouraging actions that lead to favorable outcomes and discouraging detrimental ones. This reward system is instrumental in enabling the agent to iteratively refine its strategy based on the consequences of its actions, thereby enhancing its performance over time.

\begin{figure}[!ht]
    \centering
    \includegraphics[width=\columnwidth]{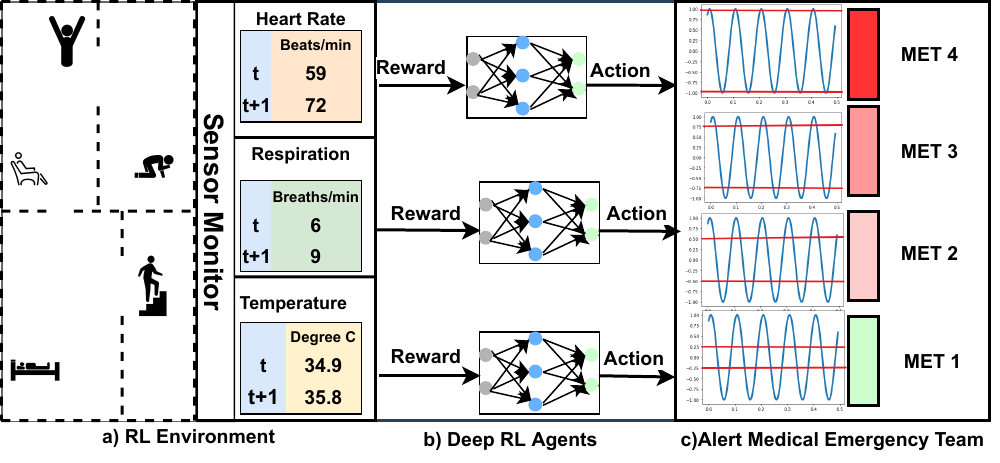}
    \caption{Human monitoring framework to monitor vital signs of the client during regular activities and alert medical emergency teams accordingly.}
    \label{fig:graph_abs}
\end{figure}

The versatility of RL has been demonstrated in various dynamic domains, such as stock market trading~\cite{ZHANG2022108543}, and is increasingly being adapted for healthcare applications, including diagnostic decisions and dynamic treatment regimes that require the consideration of delayed feedback~\cite{hong2022state}. Specifically, RL-based patient monitoring applications have focused on optimizing the timing and dosage of medications to ensure their correct administration~\cite{Watts2020,Naeem2021}. The analogy of probabilistic machine learning models, such as RL, to an ICU clinician monitoring a patient's state and making subsequent decisions based on observed changes, underscores the potential of RL in healthcare~\cite{Chen2021}.

This study addresses the challenge of monitoring multiple vital signs of the human body, tracking health status, and enabling timely interventions during emergencies by proposing an innovative approach that employs multiple deep learning agents within a healthcare monitoring environment. Each agent is responsible for monitoring specific vital signs, progressively learning threshold levels based on Modified Early Warning Scores (MEWS) and rewards accumulated from previous iterations. These well-trained deep reinforcement learning (DRL) agents are capable of monitoring parameters such as heart rate, respiration rate, and temperature, alerting clinical teams in case of deviations from predefined thresholds. While reinforcement learning (RL) has been applied in gaming environments like Deep Q-Networks (DQN) and sensor-based systems, this study offers novel contributions by integrating multi-agent reinforcement learning (MARL) within healthcare, creating a real-time, autonomous patient monitoring system. Unlike conventional RL models that focus on singular tasks, this approach uses multiple agents to monitor different physiological parameters simultaneously, enabling concurrent learning and more dynamic, comprehensive system responses. Additionally, the study introduces a novel reward mechanism that optimizes agent behavior based on MEWS, allowing agents to learn more efficiently and facilitate timely medical interventions, thus enhancing healthcare decision-making through a multi-agent, clinically informed AI framework. This represents a significant advancement in AI-driven patient monitoring.

The primary aim of this study is to learn human behavior patterns in the context of clinical health by deploying a DRL agent for each physiological feature. These agents are designed to monitor, learn, and alert the respective clinical teams if any vital signs deviate from the norms established by MEWS. We introduce a novel approach for rewarding the actions of RL agents to facilitate the learning of behavior patterns. The generic monitoring environment developed in this study supports multi-agent functionality to monitor various vital signs of a patient, thereby introducing a new paradigm for remotely monitoring patients' health status using a multi-agent DRL environment as shown in Fig.~\ref{fig:graph_abs}.

The contributions of this study are as follows:
\begin{itemize}
    \item Introduction of a novel approach for rewarding RL agents' actions to foster the learning of behavior patterns.
    \item Development of a generic monitoring environment that accommodates multi-agents for monitoring various vital signs of a patient.
    \item Establishment of a new paradigm for remotely monitoring patients' health status utilizing the multi-agent DRL environment.
\end{itemize}

The paper is organized as follows: Section~\ref{relatedworks} presents related works in the RL community, specifically focusing on learning human behavior patterns and applications in the healthcare domain. The research problem formulation and the proposed multi-agent DRL methodology are detailed in Section~\ref{methods}. Section~\ref{experiment} evaluates the proposed methodology on 10 different subject vital signs, and baseline models are discussed. In Section~\ref{results}, the results of the proposed approach are compared with baseline models, and hyper-parameter optimization of the learning rate and discount factor is discussed. Based on the results, applications of the proposed framework are discussed in Section~\ref{discussion}. Section~\ref{conclusion} concludes the paper, including limitations and future work.

\section{Related Works}\label{relatedworks}
\subsection{Machine Learning in Healthcare}
Machine learning has transformed healthcare with its ability to predict, detect, and monitor, as noted in~\cite{Rastogi2022}. Supervised learning algorithms can learn from labeled data and make predictions or classify based on the input features~\cite{mahesh2020machine}. For example, machine learning or deep learning techniques can predict human vital signs like heart rate or classify physical activities~\cite{alsareii2022physical}. In a study by Oyeleye et al.~\cite{Oyeleye2022}, machine learning and deep learning models were used to estimate heart rate using data from wearable devices. The authors tested different regression algorithms including linear regression, k-nearest neighbor, decision tree, random forest, autoregressive integrated moving average, support vector regressor, and long short-term memory recurrent neural networks. Similarly, Luo et al.~\cite{Luo2020} utilized the LSTM model to predict heart rate based on five factors: heart rate signal, gender, age, physical activities~\cite{MinhDang2020}, and mental state. Unsupervised learning algorithms learn from unlabeled data and find patterns using association and clustering techniques~\cite{sheng2020unsupervised, norgaard2018synthetic}. Sheng and Huber~\cite{sheng2020unsupervised} developed an unsupervised method with an encoder and decoder network to identify similar physical activities using clustering, which achieved a clustering accuracy of 85\% based on learning embeddings and behavior clusters. RL, on the other hand, does not require prior knowledge or information and works on an environment-driven approach~\cite{sarker2021machine}. The agents learn through receiving rewards or penalties based on their actions which is called as experience. Unlike supervised learning, RL can learn a sequence of actions through exploration and exploitation and does not require extensive labeled data for data-driven models~\cite{Lou2021}.

\subsection{Mimic Human Behavior Patterns}
Tirumala et al.~\cite{tirumala2020behavior} studied how to understand human behavior patterns and identify common movements and interactions in a set of related tasks and situations. They used probabilistic trajectory models to develop a framework for hierarchical reinforcement learning (HRL). Janssen et al.~\cite{janssen2022hierarchical} suggested breaking down a complex task such as biological behavior into smaller parts, with HRL able to organize sequential actions into a temporary option. They compared biological behavior to options in HRL. Tsiakas et al.~\cite{tsiakas2017interactive} proposed a human-centered cyber-physical systems framework for personalized human-robot collaboration and training, focusing on monitoring and evaluating human behavior. The authors aimed to effectively predict human attention with the minimum and least intrusive sensors. Kubota et al.~\cite{kubota2022methods} investigated how robots can adapt to the behavior of people with cognitive impairments for cognitive neuro-rehabilitation. They explored different types of robots for therapeutic, companion, and assistive applications. For health applications, robots must be able to perceive and understand human behavior, which includes high-level behaviors like cognitive abilities and engagement, as well as low-level behaviors like speech, gesture, and physiological signs~\cite{elouni2020intelligent}.

\subsection{RL in Healthcare}
Lisowska et al.~\cite{lisowska2021personalized} developed a digital intervention for cancer patients to promote positive health habits and lifestyle changes. They used RL to determine the best time to send intervention prompts to the patients and employed three RL approaches (Deep Q-Learning, Advance Actor Critic, and proximal policy optimization) to create a virtual coach for sending prompts. Other studies have also shown that personalized messages can increase physical activity in type 2 diabetes patients~\cite{yom2017encouraging}. Li et al.~\cite{li2022electronic} proposed a RL approach based on electronic health records for sequential decision-making tasks. They used a model-free Deep Q Networks (DQN) algorithm to make clinical decisions based on patient data and achieved better results with cooperative multi-agent RL. R decision-making can also be used for human activity recognition, as shown in a study that proposed a dynamic weight assignment network architecture and used a twin delayed deep deterministic algorithm inspired by various other RL algorithms~\cite{guo2021deep, shaik2024framu}.

Reinforcement learning (RL) has been extensively researched across various domains, including gaming, human behavior modeling, and socially assistive robotics. While these applications have demonstrated the effectiveness of RL in controlled environments, the deployment of physical robots in sensitive settings like hospitals, elderly care, and mental health facilities poses significant safety risks to patients and staff~\cite{mivseikis2020lio}. Moreover, traditional supervised and unsupervised learning methods used in health monitoring systems often struggle to handle the uncertain and dynamic nature of hospital environments. In contrast, our study introduces a novel application of multi-agent reinforcement learning (MARL) for healthcare, where virtual agents, rather than physical robots, autonomously monitor and learn from patient vital signs in real-time. This approach goes beyond the typical uses of RL by integrating multi-agent systems in a healthcare setting, enabling concurrent learning across multiple physiological parameters. Each agent adapts its behavior based on changes in vital signs and takes actions to alert clinical teams during emergencies. Our framework not only addresses the challenges of safety and uncertainty in dynamic healthcare environments but also introduces a clinically-informed reward mechanism that enhances the agents' decision-making abilities, making this application of AI a novel contribution to healthcare monitoring.




\begin{figure*}[!ht]
    \centering
    \includegraphics[width=0.8\textwidth]{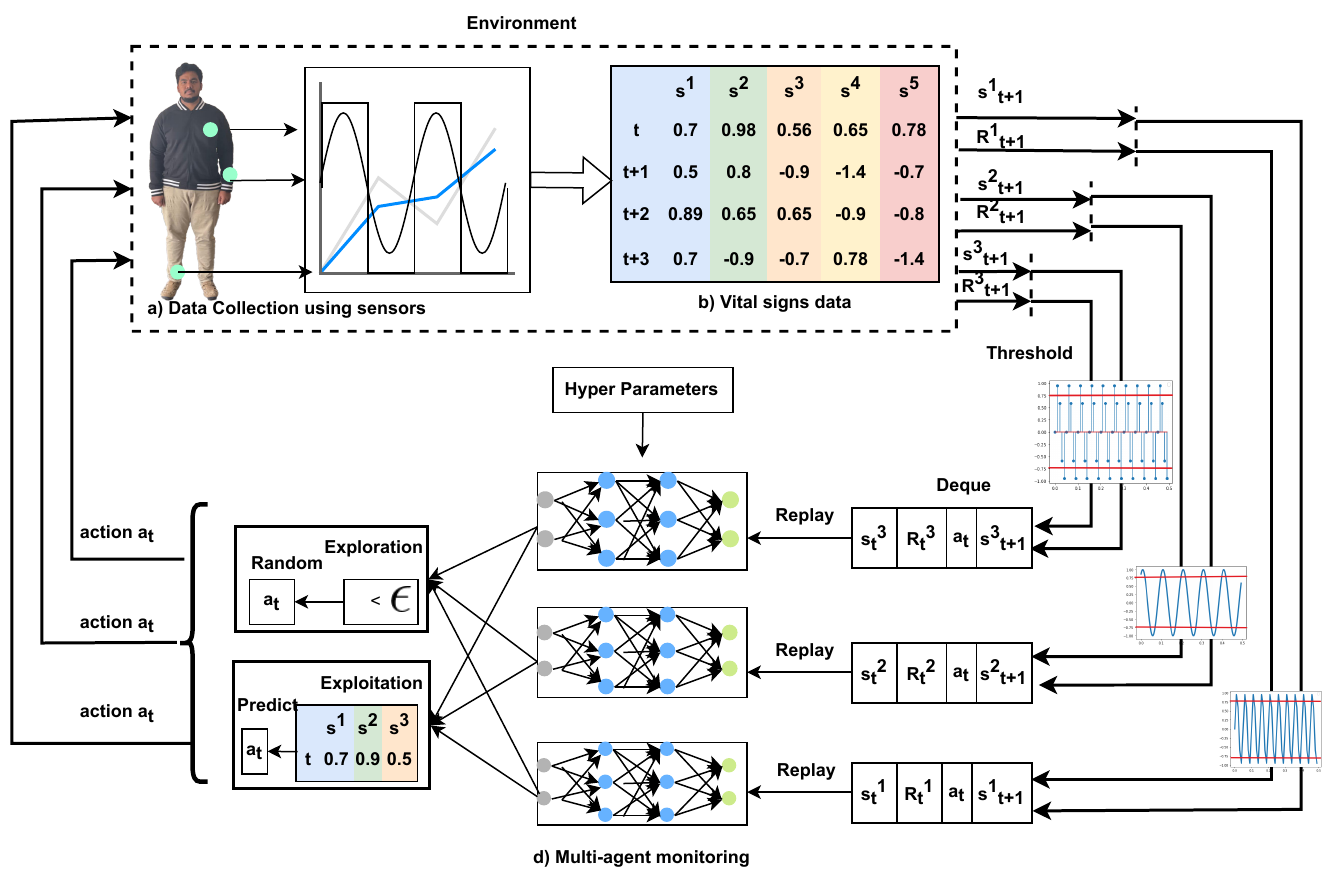}
    \caption{Multi-agent monitoring framework}
    \label{fig:multi-agent}
\end{figure*}

\section{DRL Monitoring Framework}\label{methods}
In this section, the design of a human behavior monitoring system, DRL monitoring framework, that uses R is presented in detail. The aim of the system is to monitor vital signs to learn human behavior patterns and ensure clinical safety in an uncertain environment. The proposed framework involves a multi-agent system where each vital sign state is observed by an individual agent, as shown in Fig.\ref{fig:multi-agent}. A DRL algorithm, DQN, is used to learn effective strategies in the sequential decision-making process without prior knowledge through trial-and-error interactions with the environment\cite{10.1145/3477600,shaik2023exploring}.

\subsection{Technical Background}
The challenge addressed in this research is the development of a multi-agent framework for real-time health status monitoring by learning and interpreting patterns in vital signs through wearable sensors. The agents must detect deviations from normal vital sign patterns that exceed Modified Early Warning Scores (MEWS) thresholds and alert the emergency team accordingly.

To formulate this problem, we leverage the framework of Markov Decision Processes (MDP), expressed as a 5-tuple $M = (S, A, P, R, \gamma)$. Here, $S$ represents the finite state space, where each state $s_{t} \in S$ corresponds to a distinct combination of vital sign readings at time $t$. The action set $A$ comprises potential alerts the agents can issue based on the observed vital signs. The transition function $P(s, a, s')$ models the probability of moving from state $s$ to state $s'$ upon taking action $a$, reflecting the dynamic nature of human vital signs.

Central to our approach is the reward function $R(s, a)$, which is defined to prioritize actions that lead to the early detection of potential health risks, thereby enabling timely intervention. This is mathematically represented as:

\begin{equation}\label{reward}
R(s_{t},a_{t}) = \sum_{t=0}^{\infty} \gamma^{t} r_{t},
\end{equation}

where $\gamma$ is the discount factor that balances the importance of immediate versus future rewards, ensuring the agents' actions are aligned with long-term health monitoring objectives.

The goal is to discover an optimal policy $\pi(s_{t})$ that maximizes the expected reward by selecting the most appropriate action $a_{t}$ in any given state $s_{t}$. This optimization is achieved through the iterative update of the Q-function, as outlined in the Bellman equation:

\begin{equation}\label{optimal-bellman}
Q^{new}(s_{t},a_{t}) \leftarrow (1-\alpha) Q(s_{t},a_{t}) + \alpha \left(r_{t} + \gamma \max_{a} Q(s_{t+1},a)\right),
\end{equation}

where $\alpha$ represents the learning rate, influencing the integration of new information into the Q-function. Through this process, the agents continually refine their decision-making strategy, enhancing the system's capability to monitor and respond to emerging health risks effectively.

\begin{table}[!ht]
\scriptsize
\caption{Modified Early Warning Scores~\cite{signscanberra}}
\label{tab: mews}
\centering
\resizebox{\columnwidth}{!}{%
\begin{tabular}{@{}cccccccccc@{}}
\toprule
\textbf{MEWS} &
  \cellcolor[HTML]{B2A1C7}4/MET &
  \cellcolor[HTML]{FB9E8F}3 &
  \cellcolor[HTML]{FDE9D9}2 &
  \cellcolor[HTML]{FFFFCC}1 &
  0 &
  \cellcolor[HTML]{FFFFCC}1 &
  \cellcolor[HTML]{FDE9D9}2 &
  \cellcolor[HTML]{FB9E8F}3 &
  \cellcolor[HTML]{B2A1C7}4/MET \\ \midrule
\begin{tabular}[c]{@{}c@{}}Respiratory \\ Rate\end{tabular}  & $\leq$4  & 5-8   &           &           & 9-20      & 21-24     & 25-30   & 31-35    & $\geq$36    \\ \midrule
\begin{tabular}[c]{@{}c@{}}Oxygen \\ Saturation\end{tabular} & $\leq$84 & 85-89 & 90-92     & 93-94     & $\geq$95       &           &         &          &        \\ \midrule
Temperature                                                  &     & $\leq$34.0 & 34.1-35.0 & 35.1-36.0 & 36.1-37.9 & 38.0-38.5 & $\geq$38.6   &          &        \\ \midrule
Heart Rate                                                   & $\leq$39 &       &           & 40-49     & 50-99     & 100-109   & 110-129 & 130-139  & $\geq$140   \\ \midrule
\begin{tabular}[c]{@{}c@{}}Sedation \\ Score\end{tabular}    &     &       &           &           & Awake     &           & Mild    & Moderate & Severe \\ \bottomrule
\end{tabular}}
\end{table}

\subsection{Monitoring Environment}\label{Environment}
A custom RL monitoring system based on MDP has been created to have human vital signs data serve as the observation space $S$, action space $A$ for learning agents to make decisions, and rewards $R$ for the agents' actions as depicted in Fig.~\ref{fig:multi-agent}. This study introduces a novel isolated multi-agent MDP framework that allows multi-agents to share the same environment and make decisions based on the health parameters they are monitoring, receiving rewards without being influenced by the decisions of other agents. The goal of all agents in this environment is to monitor the health of patients using the predefined MEWS, as shown in Tab.~\ref{tab: mews}. In healthcare, each vital sign plays a critical role in determining a person's clinical safety. 

In the current framework, we have implemented three RL agents to monitor heart rate, respiration, and temperature. These agents operate primarily in cooperative mode, sharing information about the patient’s health status and working together to ensure timely interventions. Cooperation allows the agents to pool rewards from collective actions, improving overall system learning. However, when multiple patients are being monitored or resource constraints arise (e.g., limited access to medical personnel), the agents may enter competitive mode. In this mode, agents prioritize the most critical health states and may compete for resources by adjusting the urgency of alerts based on the patient's condition.

As the number of agents increases, the framework is designed to scale effectively. Each additional agent monitors new physiological parameters or additional patients, with the system adjusting the reward mechanism and communication strategy to maintain efficient performance. The system remains modular, enabling easy expansion without significantly impacting computational load or decision-making speed. Importantly, the system’s ability to operate in both cooperative and competitive modes ensures flexibility, allowing it to adapt to various healthcare scenarios, including large-scale monitoring in hospitals.

\subsubsection{Observation Space}
The environment in Fig.\ref{fig:multi-agent} has a state, represented by $s_{t}^{i} \epsilon S$, where $i=0,1,2,...n$, refers to observations at time $t$. The aim is to divide the state into observations and allocate them to multi-agents. Suppose $S$ represents the state of the human body, and there are three observations, $s_{t}^{0},s_{t}^{1},s_{t}^{2} \epsilon S$, that represent different internal vital signs of the human body at time $t$. The human health status is controlled by multiple internal vital signs, each with a different threshold as shown in MEWS Tab.\ref{tab: mews}. Using a single agent to monitor all the vital signs can result in a sparse rewards challenge~\cite{wang2020deep}, where the environment might produce few useful rewards and hinders the learning of an agent. Therefore, multi-agents need to be deployed for each human to monitor the critical vital signs. The expected return $E_{\pi}$ of a policy $\pi$ in a state $s$ can be defined by state-value Eq.~\ref{modfied_value_function} in the multi-agent setting, where $i=0,1,2,3,...n$ is a finite number of observations $n$ in the state.

\begin{equation}\label{modfied_value_function}
V^\pi (s^{i}) = E_{\pi}\biggl\{\ \sum _{t=0,i=0}^{\infty,n} \gamma^{t}R(s_{t},\pi(s_{t}))| s_{0}^{0}=s\biggl\}
\end{equation}

\subsubsection{Action Space}
The action space of the monitoring environment is defined based on the MEWS~\cite{signscanberra} as shown in Tab.~\ref{tab: mews}. The table presents early warning scores of adults' vital signs with the appropriate Medical Emergency Team (MET) to contact if any escalations in the health parameters. Based on the MEWS as threshold values, the action space has been segmented to have five discrete actions to communicate the vital signs to MET–0, MET-1, MET-2, MET-3, and MET-4. Each of these actions will be taken by agents based on the current state of the vital signs they are monitoring. The expected return $E_{\pi}$ for taking an action $a$ in a state $s$ under a policy $\pi$ can be measured using the action-value function $Q _{\pi}(s,a)$ defined in Eq.~\ref{modified_action_value}.

\begin{equation}\label{modified_action_value}
Q^{\pi}(s,a)  = E_{\pi}\biggl\{\ \sum _{t=0}^{\infty} \gamma^{t}R(s_{t},a_{t},\pi(s_{t}))| s_{0}=s, a_{0}=a \biggl\}
\end{equation}

\begin{table}[ht]
\scriptsize
\caption{Rewards Policy}
\label{tab:rewards}
\centering
\begin{tabular}{@{}cccccc@{}}
\toprule
\textbf{MEWS} &
  \cellcolor[HTML]{B2A1C7}\textbf{4} &
  \cellcolor[HTML]{FB9E8F}\textbf{3} &
  \cellcolor[HTML]{FDE9D9}\textbf{2} &
  \cellcolor[HTML]{FFFFCC}\textbf{1} &
  \textbf{0} \\ \midrule
\textbf{\begin{tabular}[c]{@{}c@{}}Action 0\end{tabular}}  & -4 & -3 & -2 & -1 & 10 \\ \midrule
\textbf{\begin{tabular}[c]{@{}c@{}}Action 1\end{tabular}}  & -4 & -3 & -2 & 10 & -1 \\ \midrule
\textbf{\begin{tabular}[c]{@{}c@{}}Action 2\end{tabular}}   & -4 & -3 & 10 & -1 & -2 \\ \midrule
\textbf{\begin{tabular}[c]{@{}c@{}}Action 3\end{tabular}} & -4 & 10 & -1 & -2 & -3 \\ \midrule
\textbf{\begin{tabular}[c]{@{}c@{}}Action 4\end{tabular}}  & 10 & -3 & -2 & -1 & -4 \\ \bottomrule
\end{tabular}
\end{table}

\subsubsection{Rewards}\label{rew}
The goal of RL is to maximize cumulative rewards obtained through the actions of learning agents in an environment. In traditional RL, an agent is rewarded based on its action that leads to a transition from state $s_{t}$ to $s_{t+1}$. In this study, the objective of the learning agent is to learn patterns in human vital signs. This is achieved through the design of an effective reward policy. The reward policy, as defined in this study, is calculated using Eq.~\ref{reward_cal}. The agents are positively rewarded if they monitor vital signs in a state and take the correct action from the action space to communicate with the correct MET as defined in MEWS Tab.\ref{tab: mews}. On the other hand, if the agent takes the wrong action, it is negatively rewarded. The rewards are split into five categories for the five actions in the action space based on the MET from MEWS Tab.\ref{tab: mews}. The full rewards for each action selected by the agents are presented in Tab.\ref{tab:rewards}. The reward policy utilizes the DRL agents' desire to maximize rewards in each learning iteration, making them learn the behavior patterns. Under each category, different levels of rewards were configured. For example, an observation $s_{t}^{1} \epsilon S$ at the time $t$ is related to heart rate falling under MET–4, the rewards are shown in Eq.~\ref{rewardsequ}. 

\begin{equation}\label{reward_cal}
{R(s_{t},a_{t}) =\begin{cases}
+reward &\text{if $action = MET$}\\
-reward &\text{if $action \neq MET$}

\end{cases}}
\end{equation}

\begin{equation}\label{rewardsequ}
{R(s_{t}^{1},a_{t}) =\begin{cases}
10 &\text{if $MET=4 \& action=4$}\\
-1 &\text{if $MET=4 \& action=3$}\\
-2 &\text{if $MET=4 \& action=2$}\\
-3 &\text{if $MET=4 \& action=1$}\\
-4 &\text{if $MET=4 \& action=0$}\\
\end{cases}} 
\end{equation}

\subsection{Learning Agent}

In this study, a game learning agent DQN algorithm is employed. The DQN algorithm was first introduced by DeepMind, a subsidiary of Google, for playing Atari games. It allows the agent to play games by simply observing the screen, without any prior training or knowledge about the games. The DQN algorithm approximates the Q-Learning function using neural networks, and the learning agent is rewarded based on the neural network's prediction of the best action for the current state. For this research, the reward policy is described in more detail in Section~\ref{rew}.

\subsubsection{Function Approximation}
The neural network used in this study to estimate the Q-values for each action has three layers: an input layer, a hidden layer, and an output layer. The input layer has a node for each vital sign in a state and the output layer has a node for each action in the action space. The model is configured with a relu activation function, mean square error as the loss function, and an Adam optimizer. The model is trained on the states and their corresponding rewards and, once trained, it can predict the accumulated reward.

The learning agent takes an action $a_{t} \in A$ in a transition from state $s_{t}$ to $s_{t}^{'}$ and receives a reward $R$. In this transition, the maximum Q-function value is calculated according to Eq.~\ref{modified_action_value}, and the calculated value is discounted by a discount factor $\gamma$ to prioritize immediate rewards over future rewards. The discounted future reward is combined with the current reward to obtain the target value. The difference between the prediction from the neural network and the target value forms the loss function, which is a measure of the deviation of the predicted value from the target value and can be estimated using Eq.~\ref{loss function}. The square of the loss function penalizes the agent for large loss values.

\begin{equation}\label{loss function}
   \displaystyle{ loss = (\underbrace{R+\gamma \cdot max(Q^{\pi^{*}}(s,a))}_{target\_value}- \underbrace{Q^{\pi}(s,a)}_{predicted\_value})^{2}}
\end{equation}

\subsubsection{Memorize and Replay}
The basic neural network model has a limitation in its memory capacity and can forget previous observations as they are overwritten by new observations. To mitigate this issue, a memory array that stores the previous observations including the current state $s_{t}$, action $a_{t}$, reward $R$, and next state $s_{t}^{'}$ is used. This memory array enables the neural network to be retrained using the replay method, where a random sample of previous observations from the memory is selected for training. In this study, the neural network model was retained by using a batch size of 32 previous observations.

\subsubsection{Exploration and Exploitation}

The exploration-exploitation trade-off in RL refers to the balancing act between trying out new actions to gather information and exploiting the actions that lead to the highest rewards. This balance can be modeled mathematically using the $\epsilon$-greedy algorithm, which defines a probability $\epsilon$ of choosing a random action and a probability $1-\epsilon$ of choosing the action believed to lead to the highest reward based on the current knowledge of the action-value function $Q(s_t, a)$. The equation to determine the action taken at time $t$ is as follows:

\begin{equation}\label{eq:epsilon_greedy}
a_t = \begin{cases}
random(a_t) & \text{with probability } \epsilon \\
greedy(a_t) & \text{with probability } 1-\epsilon
\end{cases}
\end{equation}

where the greedy action is defined as:

\begin{equation}\label{eq:greedy_action}
a_t = \arg\max_a Q(s_t, a)
\end{equation}

The value of $\epsilon$ determines the level of exploration versus exploitation, with smaller values leading to more exploitation and larger values leading to more exploration. Over time, as the action-value function becomes more accurate, $\epsilon$ can be decreased to allow for more exploitation and convergence to the optimal policy.

In this study, we emphasize the importance of balancing exploration and exploitation for effective patient monitoring. Exploration allows agents to discover better monitoring strategies, while exploitation ensures timely alerts by acting on learned knowledge. Through empirical testing, we found that an exploration rate $\epsilon$ between 0.1 and 0.2 provided the optimal balance in our healthcare environment. This range ensured that agents could adapt to changing patient conditions while still providing timely and accurate interventions. In critical situations with frequent health deviations, a higher exploitation rate proved beneficial, whereas environments with fewer critical events required more exploration to discover new monitoring patterns.

\subsubsection{Hyper Parameters}
Other than the parameters defined for the neural networks, a set of hyperparameters has to supply for the RL process. They are as follows:
\begin{itemize}
    \item \textbf{episodes ($\mathcal{M}$):} This is a gaming term that means the number of times an agent has to execute the learning process. 
    \item \textbf{learning\_rate($\alpha$):} Learning rate is to determine much information neural networks learn in an iteration.
    \item  \textbf{discount\_factor($\gamma$):} Discount factor ranges from 0 to 1 to limit future rewards and focus on immediate rewards.
\end{itemize}


\begin{algorithm}[!ht]
\small
\caption{multi-agents Monitoring}\label{alg3:cap}
\begin{algorithmic}[1]
\Require{\textbf{Input:} {a set of subjects $\mathcal{C}=\{1,2,\dots,C\} $};{a set of vital signs $\mathcal{V}=\{1,2,\dots,V\}$}; {Episodes $\mathcal{M}=\{1,2,\dots,M\}$}}; 
\Ensure{\textbf{Output:} Rewards achieved by agentss in each episode.\vfill}

\State $\textbf{Initialization}:observation\_space={s_{t}^{i} \epsilon S},action\_space={a_{t}\epsilon A},reward~R$, $\gamma,\epsilon,\epsilon_{decay},\epsilon_{min},memory=\emptyset,batch\_size $
\State $Set\ monitor\_length= N$
\If{action is appropriate}
\State $R \leftarrow +reward$
\Else
\State $R \leftarrow -reward$
\EndIf

\State \textbf{Define} $model \leftarrow Neural Network Model$ 
\State $memory \leftarrow memory \cup {(s_{t}, a_{t}, R, s_{t+1})}$ 

\If {$np.random.rand < \epsilon$} \Comment{Exploration}
\State $action\_value \gets random(a_{t})$
\Else \Comment{Exploitation}
\State $action\_value \gets greedy(a_{t})$
\EndIf

\For  {episode $m \in \mathcal{M}$}
    \State $score=0$
    \For {time in range(monitor\_length)}
        \State $a_{t} \leftarrow action(s_{t})$
        \State $s_{t+1}, R, done \leftarrow step(a_{t})$
        \State $memory \leftarrow memory \cup {(s_{t}, a_{t}, R, s_{t+1})}$ 
        \State $s_{t} \leftarrow s_{t+1}$
            \If {done}
                \State $display m, score$
                \State $break$
            \EndIf
    \EndFor
    \State $replay \leftarrow batch\_size$
\EndFor
\end{algorithmic}
\end{algorithm}

Algorithm ~\ref{alg3:cap} implements the proposed multi-agent human monitoring framework. It takes as input a set of subjects $\mathcal{C}={1,2,\dots,C} $ and a set of vital signs $\mathcal{V}={1,2,\dots,V}$, along with the number of episodes $\mathcal{M}={1,2,\dots,M}$. The algorithm outputs the rewards achieved by agents in each episode. Lines 1-2 initializes all the parameters needed for monitoring the environment and learning agent. Lines 3-7 present the reward policy. Lines 8-14 present the function approximation using the neural networks model, memorize \& replay, exploration \& exploitation of the DRL agent. Lines 15-28 are nested for loops with conditional statements to check if the episode is completed or not. The outer loop is to iterate each episode while resetting the environment to initial values and score to zero. The inner loop is to iterate timesteps which denote the time of the current state and calls the methods.

The patient monitoring system operates with multiple agents, each tasked with monitoring specific vital signs such as heart rate, respiration rate, and temperature. The agents are initialized with a basic action set, which includes triggering alerts, adjusting monitoring intervals, and taking no action based on the patient’s condition. At each time step, agents receive vital sign data as input and evaluate the patient's state. Based on the current state and the agent’s policy, an action is selected. The reward function provides feedback based on the timeliness and accuracy of the action: positive rewards are given for correct, timely interventions, while penalties are applied for false alarms or missed emergencies. Over time, the agents improve their performance through continuous learning and collaboration, ensuring that vital signs are monitored comprehensively and interventions are timely.

\begin{figure}[!ht]
    \centering
    \includegraphics[width=\columnwidth]{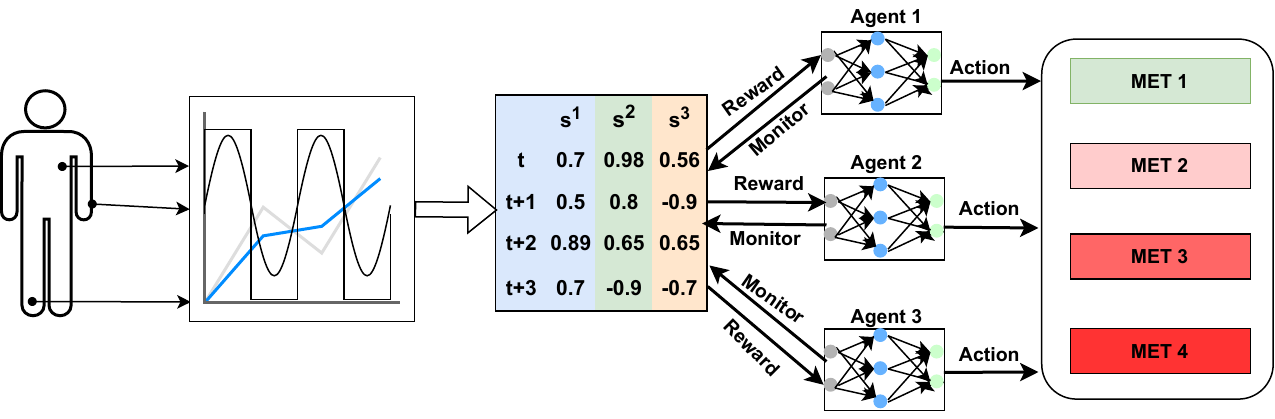}
    \caption{Experiemental Design}
    \label{fig:exp_desn}
\end{figure}

\section{Experiment}\label{experiment}
In this study, the proposed multi-agent framework was evaluated by deploying an agent for each physiological feature of a different set of subjects. The aim of the learning agents was to monitor their respective vital signs, communicate with the corresponding MET based on the estimated level of emergency, and learn the subjects' behavior patterns. All the experiments were conducted using Python programming language version 3.7.6 and related libraries such as TensorFlow, Keras, Open Gym AI, and stable\_baselines3.

\subsection{Dataset}
\begin{itemize}
\item \textbf{PPG-DaLiA~\cite{reiss2019deep}:} The dataset contains physiological and motion data of 15 subjects, recorded from both a wrist-worn device and a chest-worn device while the subjects were performing a wide range of activities under conditions close to real life.
\item \textbf{WESAD~\cite{schmidt2018introducing}:} The WESAD (Wearable Stress and Affect Detection) dataset is a collection of physiological signals recorded from participants while they perform various activities. It includes multi-modal signals such as ECG, PPG, GSR, respiration, and body temperature.
\end{itemize}

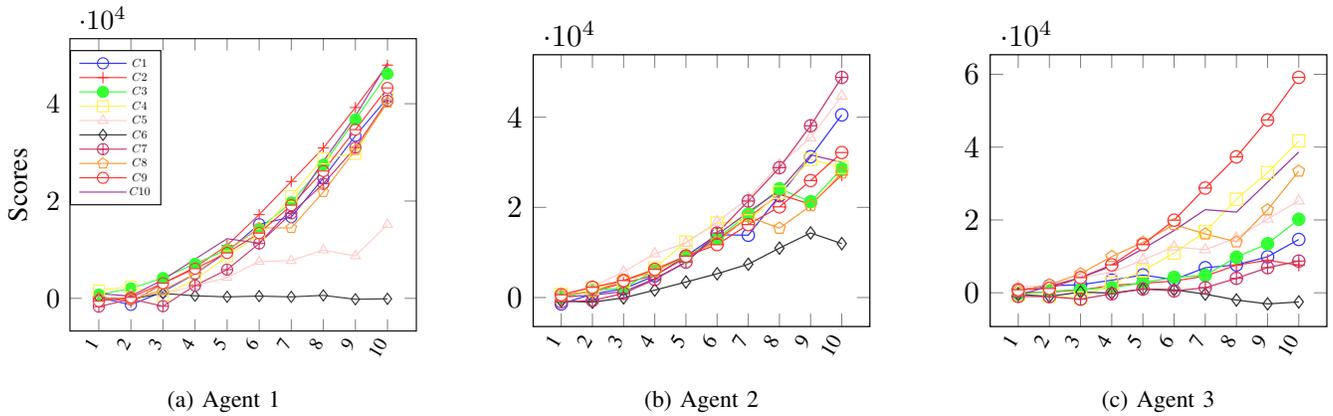
\begin{figure}[]
     \centering
     \begin{subfigure}[b]{0.45\textwidth}
        \centering
        \pgfplotstableread[col sep=comma,]{agent1.csv}\datatable
        \resizebox{\columnwidth}{!}{%
        \begin{tikzpicture}[thick,scale=0.3]
        \begin{axis}[
            width=\textwidth,
            height=5.3cm,
            xtick=data,
            xticklabels from table={\datatable}{Episode},
            x tick label style={font=\scriptsize, rotate=60, anchor=east},
            legend style={nodes={scale=0.42, transform shape},at={(0.0, 0.43)},anchor=south west},
            ylabel={Scores}]
            
            \addplot [mark=o, blue!80 ] table [x expr=\coordindex, y={C 1}]{\datatable};
            \addlegendentry{$C 1$}
            
            \addplot [mark=+, red!80 ] table [x expr=\coordindex, y={C 2}]{\datatable};
            \addlegendentry{$C 2$}
            
             \addplot [mark=*, green!80 ] table [x expr=\coordindex, y={C 3}]{\datatable};
            \addlegendentry{$C 3$}
             \addplot [mark=square, yellow!80 ] table [x expr=\coordindex, y={C 4}]{\datatable};
            \addlegendentry{$C 4$}
             \addplot [mark=triangle, pink!80 ] table [x expr=\coordindex, y={C 5}]{\datatable};
            \addlegendentry{$C 5$} 
            
            \addplot [mark=diamond, black!80 ] table [x expr=\coordindex, y={C 6}]{\datatable};
            \addlegendentry{$C 6$}
             \addplot [mark=oplus, purple!80 ] table [x expr=\coordindex, y={C 7}]{\datatable};
            \addlegendentry{$C 7$}
             \addplot [mark=pentagon, orange!80 ] table [x expr=\coordindex, y={C 8}]{\datatable};
            \addlegendentry{$C 8$}
             \addplot [mark=halfcircle, red!80 ] table [x expr=\coordindex, y={C 9}]{\datatable};
            \addlegendentry{$C 9$}
             \addplot [mark=cubes, violet!80 ] table [x expr=\coordindex, y={C 10}]{\datatable};
            \addlegendentry{$C 10$}

        \end{axis}
        \end{tikzpicture}
        }
         \caption{Agent 1}
         \label{fig:agent1_per}
     \end{subfigure}
     \vfill
     \begin{subfigure}[b]{0.45\textwidth}
         \centering
         \pgfplotstableread[col sep=comma,]{agent2.csv}\datatable
        \resizebox{\columnwidth}{!}{%
        \begin{tikzpicture}[thick,scale=0.3]
        \begin{axis}[
            width=\textwidth,
            height=4.7cm,
            xtick=data,
            xticklabels from table={\datatable}{Episode},
            x tick label style={font=\scriptsize, rotate=60, anchor=east},
            legend style={nodes={scale=0.45, transform shape},at={(0.0, 0.33)},anchor=south west}
            ,
            ylabel={Scores}
            ]
            
            \addplot [mark=o, blue!80 ] table [x expr=\coordindex, y={Subject 1}]{\datatable};
            \addlegendentry{$Subject 1$}
             \legend{}
            \addplot [mark=+, red!80 ] table [x expr=\coordindex, y={Subject 2}]{\datatable};
            \addlegendentry{$Subject 2$}
             \legend{}
             \addplot [mark=*, green!80 ] table [x expr=\coordindex, y={Subject 3}]{\datatable};
            \addlegendentry{$Subject 3$}
             \legend{}
             \addplot [mark=square, yellow!80 ] table [x expr=\coordindex, y={Subject 4}]{\datatable};
            \addlegendentry{$Subject 4$}
             \legend{}
             \addplot [mark=triangle, pink!80 ] table [x expr=\coordindex, y={Subject 5}]{\datatable};
            \addlegendentry{$Subject 5$} 
             \legend{}
            \addplot [mark=diamond, black!80 ] table [x expr=\coordindex, y={Subject 6}]{\datatable};
            \addlegendentry{$Subject 6$}
             \legend{}
             \addplot [mark=oplus, purple!80 ] table [x expr=\coordindex, y={Subject 7}]{\datatable};
            \addlegendentry{$Subject 7$}
             \legend{}
             \addplot [mark=pentagon, orange!80 ] table [x expr=\coordindex, y={Subject 8}]{\datatable};
            \addlegendentry{$Subject 8$}
             \legend{}
             \addplot [mark=halfcircle, red!80 ] table [x expr=\coordindex, y={Subject 9}]{\datatable};
            \addlegendentry{$Subject 9$}
             \legend{}
             \addplot [mark=cubes, violet!80 ] table [x expr=\coordindex, y={Subject 10}]{\datatable};
            \addlegendentry{$Subject 10$} 
             \legend{}
        
        \end{axis}
        \end{tikzpicture}
        }
         \caption{Agent 2}
         \label{fig:agent2_per}
     \end{subfigure}
     \vfill
     \begin{subfigure}[b]{0.45\textwidth}
        
         \centering
         \pgfplotstableread[col sep=comma,]{agent3.csv}\datatable
        \resizebox{\columnwidth}{!}{%
        \begin{tikzpicture}[thick,scale=0.3]
        \begin{axis}[
            width=\textwidth,
            height=4.7cm,
            xtick=data,
            xticklabels from table={\datatable}{Episode},
            x tick label style={font=\scriptsize, rotate=60, anchor=east},
            legend style={nodes={scale=0.4, transform shape,font=\footnotesize},at={(0.0, 0.43)},anchor=south west}
            ,
            ylabel={Scores}
            ]
            
            \addplot [mark=o, blue!80 ] table [x expr=\coordindex, y={Subject 1}]{\datatable};
            \addlegendentry{$Subject 1$}
             \legend{}
            \addplot [mark=+, red!80 ] table [x expr=\coordindex, y={Subject 2}]{\datatable};
            \addlegendentry{$Subject 2$}
             \legend{}
             \addplot [mark=*, green!80 ] table [x expr=\coordindex, y={Subject 3}]{\datatable};
            \addlegendentry{$Subject 3$}
             \legend{}
             \addplot [mark=square, yellow!80 ] table [x expr=\coordindex, y={Subject 4}]{\datatable};
            \addlegendentry{$Subject 4$}
             \legend{}
             \addplot [mark=triangle, pink!80 ] table [x expr=\coordindex, y={Subject 5}]{\datatable};
            \addlegendentry{$Subject 5$} 
             \legend{}
            \addplot [mark=diamond, black!80 ] table [x expr=\coordindex, y={Subject 6}]{\datatable};
            \addlegendentry{$Subject 6$}
             \legend{}
             \addplot [mark=oplus, purple!80 ] table [x expr=\coordindex, y={Subject 7}]{\datatable};
            \addlegendentry{$Subject 7$}
             \legend{}
             \addplot [mark=pentagon, orange!80 ] table [x expr=\coordindex, y={Subject 8}]{\datatable};
            \addlegendentry{$Subject 8$}
             \legend{}
             \addplot [mark=halfcircle, red!80 ] table [x expr=\coordindex, y={Subject 9}]{\datatable};
            \addlegendentry{$Subject 9$}
             \legend{}
             \addplot [mark=cubes, violet!80 ] table [x expr=\coordindex, y={Subject 10}]{\datatable};
            \addlegendentry{$Subject 10$} 
             \legend{}

        \end{axis}
        \end{tikzpicture}
        }
         \caption{Agent 3}
         \label{fig:agent3_per}
     \end{subfigure}
        \caption{DQN Agents Performance}
        \label{fig:DQNAgent}
\end{figure}

\subsection{Baseline Models}
\begin{itemize}
\item WISEML~\cite{Mallozzi2019}: Mallozzi et al. proposed an RL framework for runtime monitoring to prevent dangerous and safety-critical actions in safety-critical applications. In this framework, runtime monitoring is used to enforce properties to the agent and shape its reward during learning.
\item CA-MQL~\cite{Chen2020}: Chen et al. proposed constrained action-based MQL (CA-MQL) for UAVs to autonomously make flight decisions that consider the uncertainty of the reference point location.
\item MADDPG~\cite{lowe2017multi}: Lowe et al. introduced a deep reinforcement learning framework for multi-agent environments. This framework uses an adaptation of actor-critic methods to coordinate agents in both cooperative and competitive settings by accounting for other agents' policies. It highlights the difficulty of traditional algorithms in multi-agent scenarios and introduces policy ensembles for more robust learning.
\item QMIX~\cite{rashid2020monotonic}: Rashid et al. developed QMIX, a value-based multi-agent RL algorithm that factors joint action-values into per-agent values, allowing for decentralised policies while training in a centralised manner. QMIX demonstrated superior performance on challenging StarCraft II tasks by ensuring consistency between centralised and decentralised learning.
\item Existing RL baseline models by Li et al.~\cite{li2022electronic} were deployed to optimize sequential treatment strategies based on Electronic Health Records (EHRs) for chronic diseases using DQN. The multi-agent framework results were compared with Q-Learning and Double DQN.
\item Similarly, RL was deployed to recognize human activity using a dynamic weight assignment network architecture with TD3 (a combination of Deep Deterministic Policy Gradient (DDPG), Actor-Critic, and DQN) by Guo et al.~\cite{guo2021deep}.
\item Yom et al.~\cite{yom2017encouraging} used Advantage Actor-Critic (A2C) and Proximal Policy Optimization (PPO) algorithms to act as virtual coaches in decision-making and send personalized messages.
\end{itemize}

\subsection{Performance Measures}
In the initial phase, Cumulative Rewards were selected as the primary performance metric because they offer a direct reflection of the RL agents' success in achieving healthcare objectives. These cumulative rewards quantify the agents’ ability to make correct decisions based on real-time physiological data, which is essential for ensuring timely medical interventions. Given the critical nature of healthcare systems, focusing on cumulative rewards allowed for the evaluation of how well the agents were trained to detect early signs of health deterioration.

To provide a more holistic evaluation, we introduced additional performance metrics:
\begin{itemize}
    \item Learning Rate: This metric evaluates how quickly the agents converge to an optimal policy, which is vital in healthcare applications where rapid adaptation to changing patient conditions is crucial. Faster learning ensures that the agents can respond to emergencies in real time, improving the effectiveness of the system.
    \item Computational Complexity: This metric assesses the system’s processing demands, particularly in terms of CPU/GPU time. Minimizing computational complexity is essential in healthcare settings with resource constraints, such as hospitals or wearable monitoring devices. Lower complexity ensures that the system can run efficiently without causing delays in decision-making.
    \item Memory Usage: As the system scales to monitor multiple physiological parameters across various agents, memory usage becomes a key factor. Efficient memory utilization is critical for deploying the framework on resource-constrained devices like wearables, ensuring scalability and adaptability without compromising performance.
\end{itemize}

Incorporating these metrics provides a more comprehensive evaluation of the proposed framework, ensuring not only its effectiveness in terms of rewards but also its efficiency, scalability, and real-world deployment potential in healthcare environments.

\section{Experiment Results and Analysis}\label{results}
The advantage of RL for monitoring systems is that it can learn to handle complex, dynamic environments. Many monitoring tasks involve making decisions based on incomplete, uncertain information, and the optimal decision may depend on the context of the situation~\cite{schippers2021optimizing}. RL can learn to make decisions in these types of problems by considering the current state of the system and past experience. In this study, the aim is to leverage the RL capability to optimize the decision-making process in patient monitoring.

\begin{table}[]
\centering
\caption{Comparison of DRL and MARL Frameworks on Cumulative Rewards}
\label{tab:drl_agents_per}
\resizebox{\columnwidth}{!}{%
\begin{tabular}{@{}ccccccc@{}}
\toprule
\textbf{} &
  \multicolumn{3}{c}{\textbf{PPG-DaLia Dataset}} &
  \multicolumn{3}{c}{\textbf{WESAD Dataset}} \\ \midrule
\textbf{Method} &
  \textbf{Agent 1} &
  \textbf{Agent 2} &
  \textbf{Agent 3} &
  \textbf{Agent 1} &
  \textbf{Agent 2} &
  \textbf{Agent 3} \\ \midrule
\textbf{Q-Learning}   & 25878 & 17304 & 23688 & 25318 & 16341 & 22823 \\ \midrule
\textbf{PPO}~\cite{yom2017encouraging}          & 23688 & 20367 & 17688 & 23128 & 19404 & 16823 \\ \midrule
\textbf{A2C}~\cite{yom2017encouraging}          & 24717 & 13707 & 24369 & 24157 & 12744 & 23504 \\ \midrule
\textbf{Double DQN}~\cite{li2022electronic}   & 25569 & 15360 & 20367 & 25009 & 14397 & 19502 \\ \midrule
\textbf{DDPG}~\cite{guo2021deep}         & 26760 & 20754 & 23967 & 26200 & 19791 & 23102 \\ \midrule
\textbf{WISEML}~\cite{Mallozzi2019}       & 28654 & 25789 & 33669 & 28094 & 24826 & 32804 \\ \midrule
\textbf{CA-MQL}~\cite{Chen2020}       & 32985 & 27856 & 34685 & 32425 & 26893 & 33820 \\ \midrule
\textbf{MADDPG}~\cite{lowe2017multi}       & 42500 & 29870 & 36015 & 41200 & 28560 & 35345 \\ \midrule
\textbf{QMIX}~\cite{rashid2020monotonic}         & 44800 & 30520 & 37600 & 43200 & 29230 & 36980 \\ \midrule
\textbf{Proposed DRL} & \textbf{48354} & \textbf{30019} & \textbf{38651} & \textbf{47794} & \textbf{29056} & \textbf{37786} \\ \bottomrule
\end{tabular}}
\end{table}


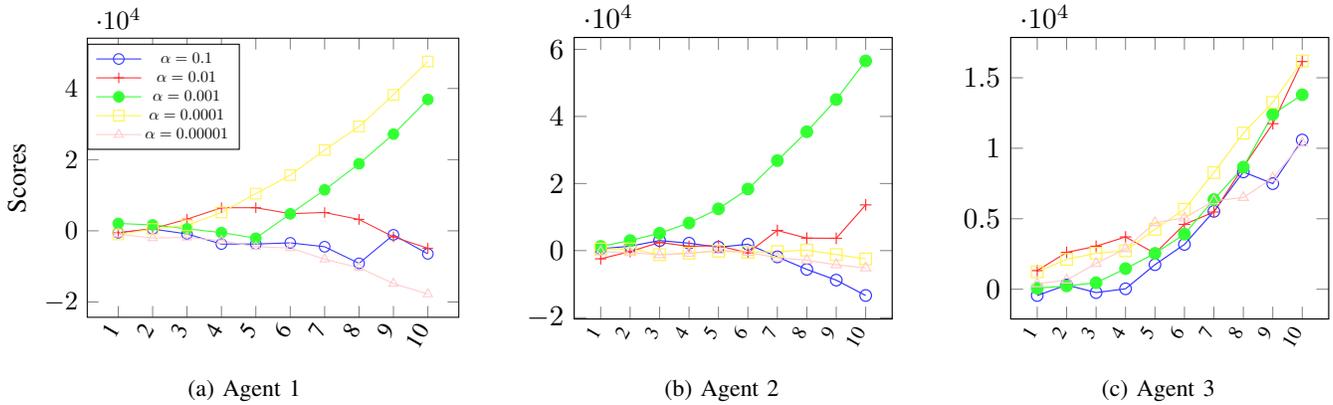
\begin{figure}[!ht]
     \centering
     \begin{subfigure}[b]{\columnwidth}
        \centering
        \pgfplotstableread[col sep=comma,]{agent1_alpha.csv}\datatable
        \resizebox{\textwidth}{!}{%
        \begin{tikzpicture}[thick,scale=0.3]
        \begin{axis}[
            width=\textwidth,
            height=5.3cm,
            xtick=data,
            xticklabels from table={\datatable}{episode},
            x tick label style={font=\small, rotate=60, anchor=east},
            legend style={nodes={scale=0.6, transform shape},at={(0.0, 0.6)},anchor=south west},
            ylabel={Scores}]
            
            \addplot [mark=o, blue!80 ] table [x expr=\coordindex, y={alpha=0.1}]{\datatable};
            \addlegendentry{$\alpha=0.1$}
            
            \addplot [mark=+, red!80 ] table [x expr=\coordindex, y={alpha=0.01}]{\datatable};
            \addlegendentry{$\alpha=0.01$}
            
             \addplot [mark=*, green!80 ] table [x expr=\coordindex, y={alpha=0.001}]{\datatable};
            \addlegendentry{$\alpha=0.001$}
             \addplot [mark=square, yellow!80 ] table [x expr=\coordindex, y={alpha=0.0001}]{\datatable};
            \addlegendentry{$\alpha=0.0001$}
             \addplot [mark=triangle, pink!80 ] table [x expr=\coordindex, y={alpha=0.00001}]{\datatable};
            \addlegendentry{$\alpha=0.00001$}

        \end{axis}
        \end{tikzpicture}
        }
         \caption{Agent 1}
         \label{fig:agent1alp}
     \end{subfigure}
     \vfill
     \begin{subfigure}[b]{\columnwidth}
         \centering
         \pgfplotstableread[col sep=comma,]{agent2_alpha.csv}\datatable
        \resizebox{\columnwidth}{!}{%
        \begin{tikzpicture}[thick,scale=0.3]
        \begin{axis}[
            width=\textwidth,
            height=5cm,
            xtick=data,
            xticklabels from table={\datatable}{episode},
            x tick label style={font=\scriptsize, rotate=60, anchor=east},
            legend style={nodes={scale=0.4, transform shape},at={(0.0, 0.43)},anchor=south west},
            ylabel={Scores}
            ]

            \addplot [mark=o, blue!80 ] table [x expr=\coordindex, y={alpha=0.1}]{\datatable};
            \addlegendentry{$\alpha=0.1$}
            \legend{}
            \addplot [mark=+, red!80 ] table [x expr=\coordindex, y={alpha=0.01}]{\datatable};
            \addlegendentry{$\alpha=0.01$}
            \legend{}
             \addplot [mark=*, green!80 ] table [x expr=\coordindex, y={alpha=0.001}]{\datatable};
            \addlegendentry{$\alpha=0.001$}
            \legend{}
             \addplot [mark=square, yellow!80 ] table [x expr=\coordindex, y={alpha=0.0001}]{\datatable};
            \addlegendentry{$\alpha=0.0001$}
            \legend{}
             \addplot [mark=triangle, pink!80 ] table [x expr=\coordindex, y={alpha=0.00001}]{\datatable};
            \addlegendentry{$\alpha=0.00001$} 
             \legend{}
        
        \end{axis}
        \end{tikzpicture}
        }
         \caption{Agent 2}
         \label{fig:agent2alp}
     \end{subfigure}
     \vfill
     \begin{subfigure}[b]{\columnwidth}
        
         \centering
         \pgfplotstableread[col sep=comma,]{agent3_alpha.csv}\datatable
        \resizebox{\columnwidth}{!}{%
        \begin{tikzpicture}[thick,scale=0.3]
        \begin{axis}[
            width=\textwidth,
            height=5cm,
            xtick=data,
            xticklabels from table={\datatable}{episode},
            x tick label style={font=\scriptsize, rotate=60, anchor=east},
            legend style={nodes={scale=0.4, transform shape,font=\footnotesize},at={(0.0, 0.43)},anchor=south west}
            ,
            ylabel={Scores}
            ]

            \addplot [mark=o, blue!80 ] table [x expr=\coordindex, y={alpha=0.1}]{\datatable};
            \addlegendentry{$\alpha=0.1$}
            
            \addplot [mark=+, red!80 ] table [x expr=\coordindex, y={alpha=0.01}]{\datatable};
            \addlegendentry{$\alpha=0.01$}
            
             \addplot [mark=*, green!80 ] table [x expr=\coordindex, y={alpha=0.001}]{\datatable};
            \addlegendentry{$\alpha=0.001$}
             \addplot [mark=square, yellow!80 ] table [x expr=\coordindex, y={alpha=0.0001}]{\datatable};
            \addlegendentry{$\alpha=0.0001$}
             \addplot [mark=triangle, pink!80 ] table [x expr=\coordindex, y={alpha=0.00001}]{\datatable};
            \addlegendentry{$\alpha=0.00001$} 
             \legend{}

        \end{axis}
        \end{tikzpicture}
        }
         \caption{Agent 3}
         \label{fig:agent3alp}
     \end{subfigure}
        \caption{Hyper Parameters - $\alpha$ optimization}
        \label{fig:hyperoptialpha}
\end{figure}

\begin{figure}[ht]
     \begin{subfigure}[b]{\columnwidth}
        \centering
        \pgfplotstableread[col sep=comma,]{agent1_gamma.csv}\datatable
        \resizebox{\columnwidth}{!}{%
        \begin{tikzpicture}[thick,scale=0.3]
        \begin{axis}[
            width=\textwidth,
            height=4.9cm,
            xtick=data,
            xticklabels from table={\datatable}{episode},
            x tick label style={font=\scriptsize, rotate=60, anchor=east},
            legend style={nodes={scale=0.6, transform shape},at={(0.0, 0.58)},anchor=south west},
            ylabel={Scores}]
            
            \addplot [mark=o, blue!80 ] table [x expr=\coordindex, y={gamma=0.95}]{\datatable};
            \addlegendentry{$\gamma=0.95$}
            
            \addplot [mark=+, red!80 ] table [x expr=\coordindex, y={gamma=0.9}]{\datatable};
            \addlegendentry{$\gamma=0.9$}
            
             \addplot [mark=*, green!80 ] table [x expr=\coordindex, y={gamma=0.85}]{\datatable};
            \addlegendentry{$\gamma=0.85$}
             \addplot [mark=square, yellow!80 ] table [x expr=\coordindex, y={gamma=0.80}]{\datatable};
            \addlegendentry{$\gamma=0.80$}
             \addplot [mark=triangle, pink!80 ] table [x expr=\coordindex, y={gamma=0.75}]{\datatable};
            \addlegendentry{$\gamma=0.75$}

        \end{axis}
        \end{tikzpicture}
        }
         \caption{Agent 1}
         \label{fig:agent1gam}
     \end{subfigure}%
     \vfill
     \begin{subfigure}[b]{\columnwidth}
         \centering
         \pgfplotstableread[col sep=comma,]{agent2_gamma.csv}\datatable
        \resizebox{\columnwidth}{!}{%
        \begin{tikzpicture}[thick,scale=0.3]
        \begin{axis}[
            width=\textwidth,
            height=4.5cm,
            xtick=data,
            xticklabels from table={\datatable}{episode},
            x tick label style={font=\scriptsize, rotate=60, anchor=east},
            legend style={nodes={scale=0.6, transform shape},at={(0.0, 0.43)},anchor=south west}
            ,
            ylabel={Scores}
            ]
            \addplot [mark=o, blue!80 ] table [x expr=\coordindex, y={gamma=0.95}]{\datatable};
            \addlegendentry{$\gamma=0.95$}
             \legend{}            
            \addplot [mark=+, red!80 ] table [x expr=\coordindex, y={gamma=0.9}]{\datatable};
            \addlegendentry{$\gamma=0.9$}
             \legend{}            
             \addplot [mark=*, green!80 ] table [x expr=\coordindex, y={gamma=0.85}]{\datatable};
            \addlegendentry{$\gamma=0.85$}
             \legend{}            
             \addplot [mark=square, yellow!80 ] table [x expr=\coordindex, y={gamma=0.80}]{\datatable};
            \addlegendentry{$\gamma=0.80$}
             \legend{}
             \addplot [mark=triangle, pink!80 ] table [x expr=\coordindex, y={gamma=0.75}]{\datatable};
            \addlegendentry{$\gamma=0.75$} 
             \legend{}
        
        \end{axis}
        \end{tikzpicture}
        }
         \caption{Agent 2}
         \label{fig:agent2gam}
     \end{subfigure}
     \vfill
     \begin{subfigure}[b]{\columnwidth}
        
         \centering
         \pgfplotstableread[col sep=comma,]{agent3_gamma.csv}\datatable
        \resizebox{\columnwidth}{!}{%
        \begin{tikzpicture}[thick,scale=0.3]
        \begin{axis}[
            width=\textwidth,
            height=4.5cm,
            xtick=data,
            xticklabels from table={\datatable}{episode},
            x tick label style={font=\scriptsize, rotate=60, anchor=east},
            legend style={nodes={scale=0.6, transform shape,font=\footnotesize},at={(0.0, 0.43)},anchor=south west}
            ,
            ylabel={Scores}
            ]
            \addplot [mark=o, blue!80 ] table [x expr=\coordindex, y={gamma=0.95}]{\datatable};
            \addlegendentry{$\gamma=0.95$}
             \legend{}            
            \addplot [mark=+, red!80 ] table [x expr=\coordindex, y={gamma=0.9}]{\datatable};
            \addlegendentry{$\gamma=0.9$}
             \legend{}            
             \addplot [mark=*, green!80 ] table [x expr=\coordindex, y={gamma=0.85}]{\datatable};
            \addlegendentry{$\gamma=0.85$}
             \legend{}
             \addplot [mark=square, yellow!80 ] table [x expr=\coordindex, y={gamma=0.80}]{\datatable};
            \addlegendentry{$\gamma=0.80$}
             \legend{}
             \addplot [mark=triangle, pink!80 ] table [x expr=\coordindex, y={gamma=0.75}]{\datatable};
            \addlegendentry{$\gamma=0.75$} 
             \legend{}
        \end{axis}
        \end{tikzpicture}
        }
         \caption{Agent 3}
         \label{fig:agent3gam}
     \end{subfigure}
        \caption{Hyper Parameters - $\gamma$ optimization}
        \label{fig:hyperoptigamma}
\end{figure}
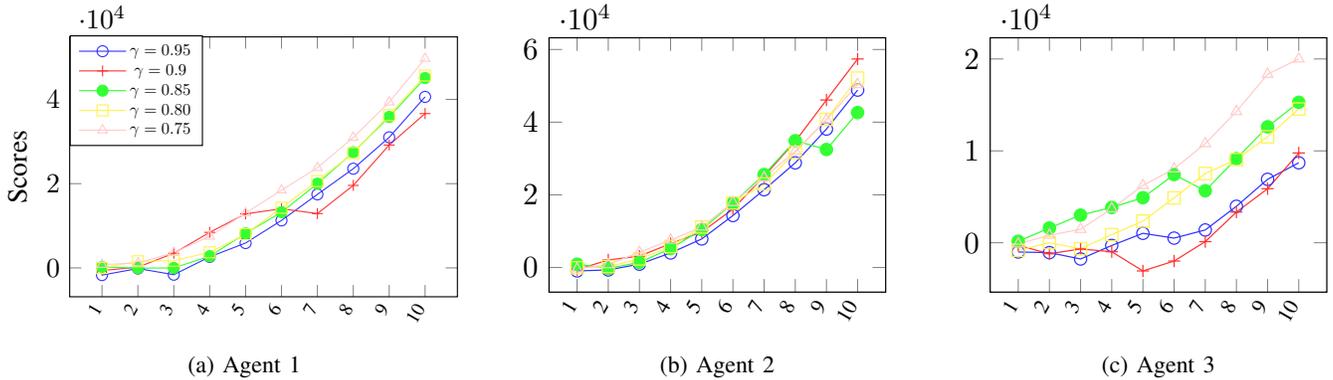
\subsection{DRL Agents Performance}
The performance of the proposed DRL framework was evaluated using two datasets, with a focus on cumulative rewards, learning rate, computational complexity, and memory usage. Additionally, we expanded our comparison to include multi-agent RL frameworks, MADDPG and QMIX, to assess how well these frameworks handle the complexities of real-time health monitoring tasks.

The results are summarized in Tab.~\ref{tab:drl_agents_per}, which includes the performance of single-agent RL methods (Q-Learning, PPO, A2C, and DDPG) and multi-agent RL models (MADDPG and QMIX). The proposed DRL framework consistently outperforms all other models in terms of cumulative rewards, with significant improvements over the baseline methods. 

As shown in Tab.~\ref{tab:drl_agents_per}, the proposed DRL framework surpasses both MADDPG and QMIX in cumulative rewards for both datasets, particularly excelling in agent 1’s performance on the PPG-DaLia dataset. This indicates that our framework's design, which includes a tailored reward mechanism based on Modified Early Warning Scores (MEWS), enables more efficient learning in healthcare environments. Additionally, the exploration-exploitation trade-off in our system is better optimized for the variability of physiological data.

Beyond cumulative rewards, we evaluated the proposed DRL framework against baseline models using additional performance metrics, including learning rate, computational complexity, and memory usage, as shown in Tab.~\ref{tab:additional_metrics}. The proposed DRL framework showed superior performance across all these metrics, indicating its suitability for real-time applications in resource-constrained healthcare environments.

In terms of learning rate, the proposed DRL framework converged after 850 epochs, outperforming all baseline models, including Q-Learning (1200 epochs) and Double DQN (1100 epochs). This faster convergence demonstrates the DRL framework’s enhanced efficiency in learning complex healthcare scenarios. Faster learning is especially critical in healthcare, where timely interventions directly impact patient outcomes. The use of multiple agents, each dedicated to a specific physiological metric, accelerates policy optimization and enhances responsiveness in dynamic, real-world environments.

For computational complexity, the proposed DRL framework exhibited a significantly lower iteration time of 0.70 seconds, outperforming more complex multi-agent models like CA-MQL (1.30 seconds) and PPO (1.10 seconds). This indicates that the framework is computationally efficient, making it ideal for real-time healthcare monitoring where decision delays could compromise patient safety. This improved efficiency is due to an optimized reward structure and action space, which reduces the time required for decision-making without compromising accuracy.

In terms of memory usage, the DRL framework consumed 110MB, which is lower than all other baseline models, such as DDPG (160MB) and CA-MQL (175MB). This low memory footprint is crucial for deploying the framework on resource-constrained hardware like wearable devices or low-power hospital systems. The efficient memory usage ensures the system can scale with additional agents without overloading system resources, making the framework suitable for large-scale healthcare applications.

\begin{table}[h!]
\centering
\caption{Evaluation of DRL Framework and Baseline Models on Additional Metrics}
\label{tab:additional_metrics}
\resizebox{\columnwidth}{!}{%
\begin{tabular}{@{}lccc@{}}
\toprule
\textbf{RL Method}         & \makecell{\textbf{Learning Rate} \\ \textbf{(Epochs to Converge)}} & \makecell{\textbf{Computational Complexity} \\ \textbf{(Time in Seconds)}} & \makecell{\textbf{Memory Usage} \\ \textbf{(MB)}} \\ \midrule
\textbf{Q-Learning}        & 1200                                       & 0.85s per iteration                                & 120MB                      \\ \midrule
\textbf{PPO}~\cite{yom2017encouraging}                & 900                                        & 1.10s per iteration                                & 150MB                      \\ \midrule
\textbf{A2C}~\cite{yom2017encouraging}              & 1000                                       & 1.05s per iteration                                & 140MB                      \\ \midrule
\textbf{Double DQN}~\cite{li2022electronic}      & 1100                                       & 0.95s per iteration                                & 135MB                      \\ \midrule
\textbf{DDPG}~\cite{guo2021deep}               & 950                                        & 1.20s per iteration                                & 160MB                      \\ \midrule
\textbf{WISEML}~\cite{Mallozzi2019}             & 900                                        & 1.15s per iteration                                & 145MB                      \\ \midrule
\textbf{CA-MQL}~\cite{Chen2020}           & 1000                                       & 1.30s per iteration                                & 175MB                      \\ \midrule
\textbf{MADDPG}~\cite{lowe2017multi}            & 950                                        & 1.25s per iteration                                & 155MB                      \\ \midrule
\textbf{QMIX}~\cite{rashid2020monotonic}             & 1100                                       & 1.20s per iteration                                & 165MB                      \\ \midrule
\makecell{\textbf{Proposed DRL}} & \textbf{850}                               & \textbf{0.70s per iteration}                        & \textbf{110MB}             \\ \bottomrule
\end{tabular}%
}
\end{table}

All three learning agents were fed with physiological features such as heart rate, respiration, and temperature, respectively, from the PPG-DaLiA dataset. Based on the observation space, action space, and reward policy defined for a customized gym environment for human behavior monitoring, the learning agents were run for 10 episodes, as shown in Fig.~\ref{fig:DQNAgent}. In the results, agent 1 refers to the heart rate monitoring agent, which showed a constant increase in scores for each episode for most of the subjects except subjects 5 and 6. The intermittent low scores in agent 1 performance are due to the exploration rate in DQN learning, where the algorithm tries exploring all the actions randomly instead of relying on neural networks' predictions. Similarly, agent 2 and agent 3 monitor two other physiological features, respiration and temperature, respectively. agent 2 performed better than the other two agents and achieved consistent scores for all subjects. Out of all agents, agent 3, temperature monitoring performance, was poor. This issue was traced back to the data level, where the units of the temperature thresholds in the MEWS table and the input body temperature data from the dataset were different. Still, agent 3 achieved high scores in monitoring subjects 9, 8, 4, and 10.

The reward policy designed in the proposed multi-agent framework enables agents to learn the human physiological feature patterns. For example, if a subject's heart rate is 139 beats per minute, agent 1 takes Action 3 to communicate the message to MET-3. The agent will get rewarded with +10 points only if Action 3 is taken; otherwise, the agent gets negatively rewarded according to the reward policy (Table~\ref{tab:rewards}). With this example, the results in Fig.~\ref{fig:DQNAgent} can be interpreted better. An increase in scores episode by episode, with the exception of the exploration rate, actually infers an increase in the learning curve of the agents in terms of human physiological patterns.

\subsection{Hyper-Parameters Optimization}
The DRL agents were further evaluated by hyperparameters optimization. Out of all the hyperparameters discussed in this study, two hyperparameters, learning rate ($\alpha$) and discount factor ($\gamma$), were optimized for all three agents, and the results are shown in Figs.\ref{fig:hyperoptialpha} and\ref{fig:hyperoptigamma}. The learning rate determines how much information neural networks learn in an iteration to predict action and approximate the rewards. The discount factor measures how much RL agents focus on future rewards relative to those in the immediate rewards. In Fig.\ref{fig:hyperoptialpha}, Figs.\ref{fig:agent1alp},\ref{fig:agent2alp}, and\ref{fig:agent3alp} show the agents' performance while optimizing $\alpha$ of neural networks. The x-axis of the plots represents scores (cumulative rewards) achieved by an agent in each episode shown on the y-axis. The bar plots show that the learning rate $\alpha=0.01$ is a more optimized value in all the monitoring agents. Similarly, Figs.\ref{fig:agent1gam},\ref{fig:agent2gam}, and~\ref{fig:agent3gam} present the $\gamma$ optimization of agent 1, agent 2, and agent 3, respectively. The discount factors $\gamma=0.9$ and $\gamma=0.75$ are the more optimized values for agents 2 and 3, respectively, after 10 episodes of training.

\section{Discussion}\label{discussion}
This study introduces an innovative approach to patient monitoring within the unpredictable environment of healthcare settings, employing adaptive multi-agent deep reinforcement learning (DRL) to ensure timely healthcare interventions. The fluctuating nature of vital signs, crucial indicators of patient health, necessitates a robust system capable of real-time analysis and decision-making. By leveraging the sequential decision-making prowess of RL algorithms, we have established a framework where each vital sign is monitored by a dedicated DRL agent. These agents operate within a cohesive monitoring environment, guided by meticulously defined reward policies to identify and respond to potential health emergencies based on MEWS and MET standards.

A notable aspect of our research is the emphasis on the design of the observation space for each DRL agent. This design is pivotal in ensuring the accuracy and effectiveness of the learning process, as it directly impacts the agent's ability to interpret vital sign data and make informed decisions. The challenge encountered with DRL agent 3, responsible for monitoring body temperature, underscores the importance of data consistency and the need for a harmonized observation space. The discrepancy between the temperature units in the MEWS table and the dataset highlighted a critical area for improvement, emphasizing the need for standardized data inputs to enhance agent performance.

The autonomous decision-making capability inherent in RL represents a significant advancement in supporting clinicians. By providing real-time updates on patient health, the DRL framework facilitates a proactive approach to patient care, extending its applicability beyond hospital settings to include home and specialized care environments. This adaptability is further enhanced by the strategic optimization of hyperparameters, which fine-tunes the learning process of DRL agents to achieve optimal performance. Our investigation into hyperparameters such as the learning rate and discount rate reveals the critical balance between immediate and future rewards, a balance that is essential for the effective monitoring of patient health.

Comparatively, traditional supervised learning algorithms, while accurate in predicting vital signs, fall short in dynamic healthcare environments due to their reliance on extensive labeled datasets and external supervision. The DRL approach, free from the constraints of labeled data, offers a more flexible and efficient solution for patient monitoring. However, it is essential to acknowledge the considerable effort required in data preparation and model tuning within supervised learning frameworks, which, despite their limitations, contribute significantly to the development of informed clinical decisions.

The adaptive multi-agent DRL framework proposed in this study represents a paradigm shift in patient monitoring, offering a dynamic, efficient, and scalable solution for timely healthcare interventions~\cite{shaik2024graph, shaik2023ai}. The challenges and insights gleaned from this research pave the way for future advancements in the field, promising to enhance the quality of patient care through innovative technological solutions.

\section{Conclusion}\label{conclusion}
This study has pioneered an adaptive framework for healthcare interventions using multi-agent DRL to dynamically monitor vital signs, establishing a novel approach in patient care. Through the development of a generic monitoring environment coupled with a strategic reward policy, the DRL agents were empowered to learn from and adapt to vital sign fluctuations, enabling timely interventions by healthcare professionals. Despite its innovative contributions, the research faced challenges, such as discrepancies in body temperature data scales and the absence of predictive capabilities for future vital sign trends, which limited the effectiveness of one DRL agent and the overall predictive potential of the system. Addressing these limitations, future research will focus on enhancing the framework with predictive analytics, allowing DRL agents to forecast vital sign trends and thus revolutionize patient monitoring. This advancement aims to facilitate proactive healthcare measures, significantly reducing the risk of critical health episodes and heralding a new era in adaptive patient monitoring and healthcare management. Having said that, the future direction of our research will be focused on extending the scope of the research to predict future vital signs using multi-agent DRL.

\bibliographystyle{IEEEtran}
\bibliography{bare_jrnl_new_sample4}






\vfill

\end{document}